\documentclass{ifacconf}

\usepackage{graphicx}      
\usepackage{natbib}        
\usepackage{amsmath}
\usepackage{float}
\usepackage{xcolor}
\usepackage{amssymb}
\usepackage{subcaption}
\usepackage{ulem}
\begin{document}
\begin{frontmatter}

\title{An Integrated Visual Servoing Framework for Precise Robotic Pruning Operations in Modern Commercial Orchard} 

\thanks[footnoteinfo]{© 2025 the authors. This work has been accepted to IFAC for publication under a Creative Commons Licence CC-BY-NC-ND.}

\author[First]{Dawood Ahmed} 
\author[Second]{Basit Muhammad Imran} 
\author[Third]{Martin Churuvija}
\author[Third,Fifth]{Manoj Karkee}

\address[First]{Cornell University, 
   Ithaca, NY 14850 USA (e-mail: da542@cornell.edu).}
\address[Second]{Virginia Tech, 
   Blacksburg, VA 24060 USA (e-mail: basit@vt.edu)}
\address[Third]{Washington State University, 
   Prosser, WA 99350 USA (e-mail: martin.churuvija@wsu.edu)}
\address[Fourth]{Cornell University, 
   Ithaca, NY 14850 USA (e-mail: mk2684@cornell.edu)}
\address[Fifth]{Washington State  University, 
   Prosser, WA 99350 USA (e-mail: manoj.karkee@wsu.edu)}

\begin{abstract}                
This study presents a vision-guided robotic control system for automated fruit tree pruning applications. Traditional pruning practices are labor-intensive and limit agricultural efficiency and scalability, highlighting the need for advanced automation. A key challenge is the precise, robust positioning of the cutting tool in complex orchard environments, where dense branches and occlusions make target access difficult. To address this, an Intel RealSense D435 camera is mounted on the flange of a UR5e robotic arm and CoTracker3, a transformer-based point tracker, is utilized for visual servoing control that centers tracked points in the camera view. The system integrates proportional control with iterative inverse kinematics to achieve precise end-effector positioning. The system was validated in Gazebo simulation, achieving a 77.77\% success rate within 5mm positional tolerance and 100\% success rate within 10mm tolerance, with a mean end-effector error of 4.28 ± 1.36 mm.  The vision controller demonstrated robust performance across diverse target positions within the pixel workspace. The results validate the effectiveness of integrating vision-based tracking with kinematic control for precision agricultural tasks. Future work will focus on real-world implementation and the integration of force sensing for actual cutting operations.
\end{abstract}


\end{frontmatter}

\section{Introduction}
The rapidly growing population and urbanization are driving a sharp increase in food demand. Agriculture remains essential for producing fruits and crops to meet this need. However, modern agricultural practices still rely heavily on human labor, limiting crop yield, scalability, and efficiency. Pruning, a vital agricultural task, improves plant health, fruit production, and overall quality but remains labor-intensive, requiring expertise and significant manual effort \citep{zhao2016review}. With a severe shortage of skilled labor and high costs associated with manual pruning, automation has become a key research focus \citep{fimiani2023sensorless}. Recent efforts have explored robotic pruning systems with advanced vision and control algorithms, offering a sustainable, precision-based alternative to manual labor.

Despite significant progress in many areas of agricultural robotics (e.g., harvesting, thinning, and crop scouting), pruning still remains a formidable challenge. First, fruit trees exhibit highly unstructured, dynamic environments with complex branch geometries and frequent occlusions by other branches or orchard infrastructure. Second, precise manipulation is required to position the end-effector at the correct pruning poses while avoiding collisions and excessive contact forces. Therefore, there is a pressing need for accurate perception, robust point tracking, and agile control schemes to enhance the capability and practical adaptability of robotic pruning systems.

\begin{figure}
    \centering
    \includegraphics[width=1.0\linewidth]{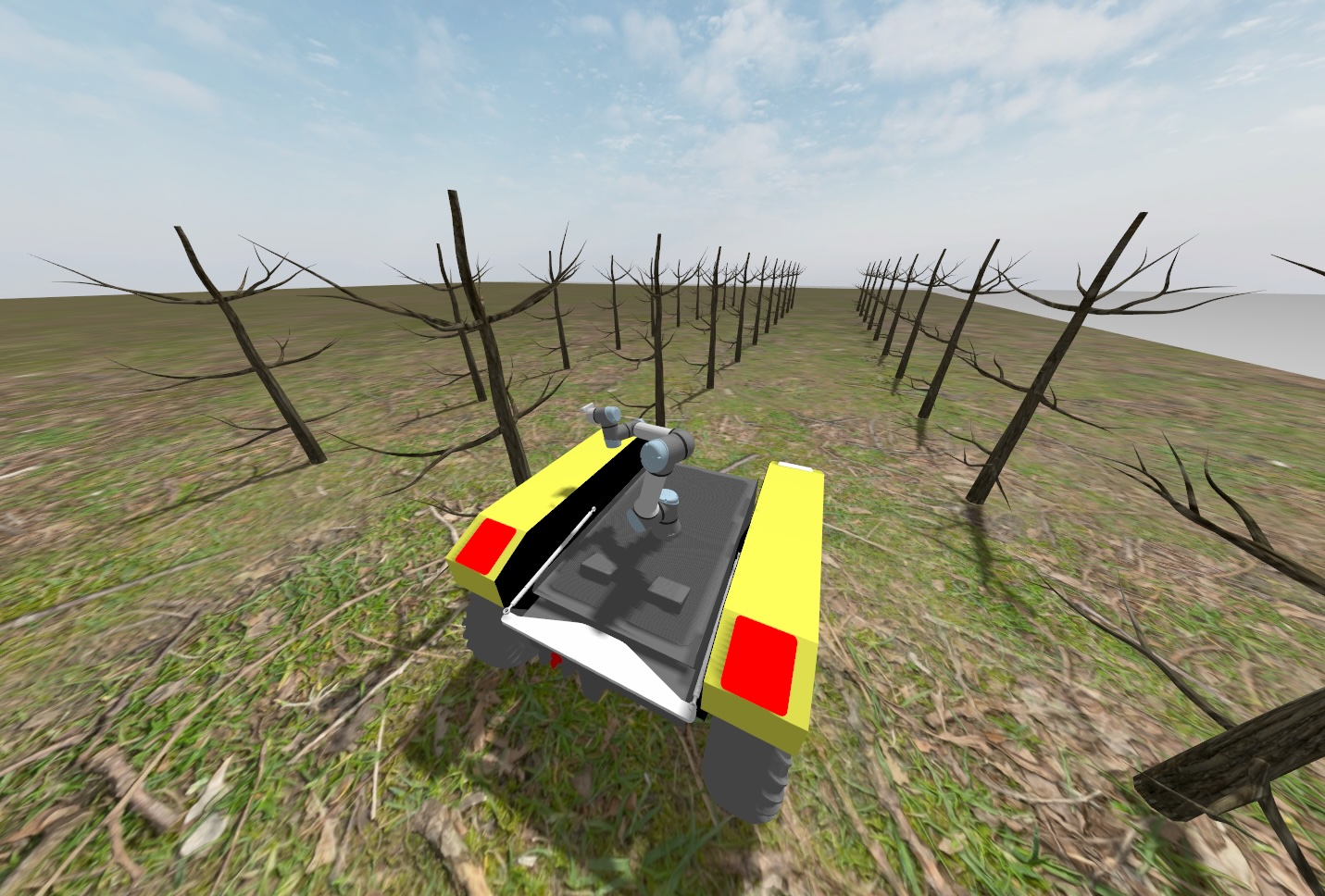}
    \caption{Simulation environment showing the Gazebo setup with the dormant trees and a Warthog bot with a mounted UR5e arm.}
    \label{fig:1}
\end{figure}

Automated pruning systems require advanced sensing, modeling, and control to function effectively in dynamic orchard environments. Previous research demonstrated that accurate branch diameter estimation \citep{ahmed2025estimating}, along with bud detection and counting \citep{Ahmed2024-vz}, provides essential data for determining optimal crop loads. These measurements inform data-driven pruning rules to identify branches and pruning locations while balancing tree structure maintenance and fruit production. After selecting pruning points, a precise, collision-free trajectory must be planned to guide the end-effector while avoiding obstacles such as trunks, wires, and posts. However, traditional open-loop or purely position-based methods can accumulate significant errors in real-world conditions, reducing accuracy and precision \citep{you2020efficient, you2023semiautonomous}.

Recent advancements in visual servoing address operational challenges in robotic pruning through closed-loop control based on camera feedback \citep{shamshiri2023review, dong2015position}. While position-based visual servoing (PBVS) relies on 3D pose estimation, image-based approaches (IBVS) minimize 2D feature errors, offering greater robustness in unstructured orchards where branches, foliage, and infrastructure complicate calibration. Yandun et al. \citep{yandun2021reaching} used deep reinforcement learning (DRL) trained on 3D vine models to navigate cluttered canopies, demonstrating adaptability but requiring extensive training. \cite{you2022precision} improved this by introducing a hybrid vision/force control framework, where vision-based policies guided the end-effector to sub-centimeter accuracy, with force feedback mitigating excessive contact with rigid branches. \cite{gebrayel2024visual} refined PBVS using iterative closest point (ICP) variants for real-time vine alignment but struggled in highly occluded scenes. These limitations highlight the need for robust visual tracking methods that maintain accuracy despite dense foliage and occlusions.

To overcome orchard challenges such as occlusions, wind-induced movements \citep{spatz2013oscillation}, and variable lighting, tracking methods have advanced beyond keypoint detection and optical flow. CoTracker3 \citep{karaev2024cotracker3} addresses these limitations with a streamlined architecture that replaces heavy correlation processing with lightweight MLPs, achieving 27\% faster performance than LoCoTrack while maintaining accuracy. Its pseudo-labeling approach allows training on real-world videos without manual annotations, improving generalization and reducing data requirements. These features make CoTracker3 well-suited for dynamic orchard environments requiring both precision and efficiency.

This research contributes to the development of an autonomous robotic system for orchard pruning in tree fruit crops (e.g., apple and cherry), with a focus on the vision-based tracking and control strategy required to reach pruning points in complex scenes. While full navigation, structural analysis and collision avoidance are assumed to be externally provided or preprocessed, this work addresses how to effectively track predetermined pruning points and control the robotic arm to achieve precise positioning. The system employs a transformer-based model to robustly track designated pruning points and guides a UR5e robotic arm to these locations. To advance autonomous pruning, this work presents a control strategy that integrates CoTracker3's robust point-tracking with iterative inverse kinematics and proportional control. The controller focuses on reaching target points using visual feedback, with depth monitoring providing a simple stopping mechanism when approaching within 20 cm of the target. Operating primarily in the 2D image plane, the system continuously updates motion based on tracked point positions, ensuring precise and responsive control. The approach is validated in a Gazebo simulation using a UR5e robotic arm (Universal Robots, Denmark) and an Intel RealSense D435 camera (Intel, California, US).

This paper is organized as follows. Section \ref{section2} outlines methodological details including a brief description of CoTracker3. Section \ref{section3} presents key results and provides a comprehensive discussion. Finally, Section \ref{section4} provides a brief conclusion along with some remarks on future work.

\section{Methodology}\label{section2}
\begin{figure*}
    \centering
    \includegraphics[width=0.99\linewidth]{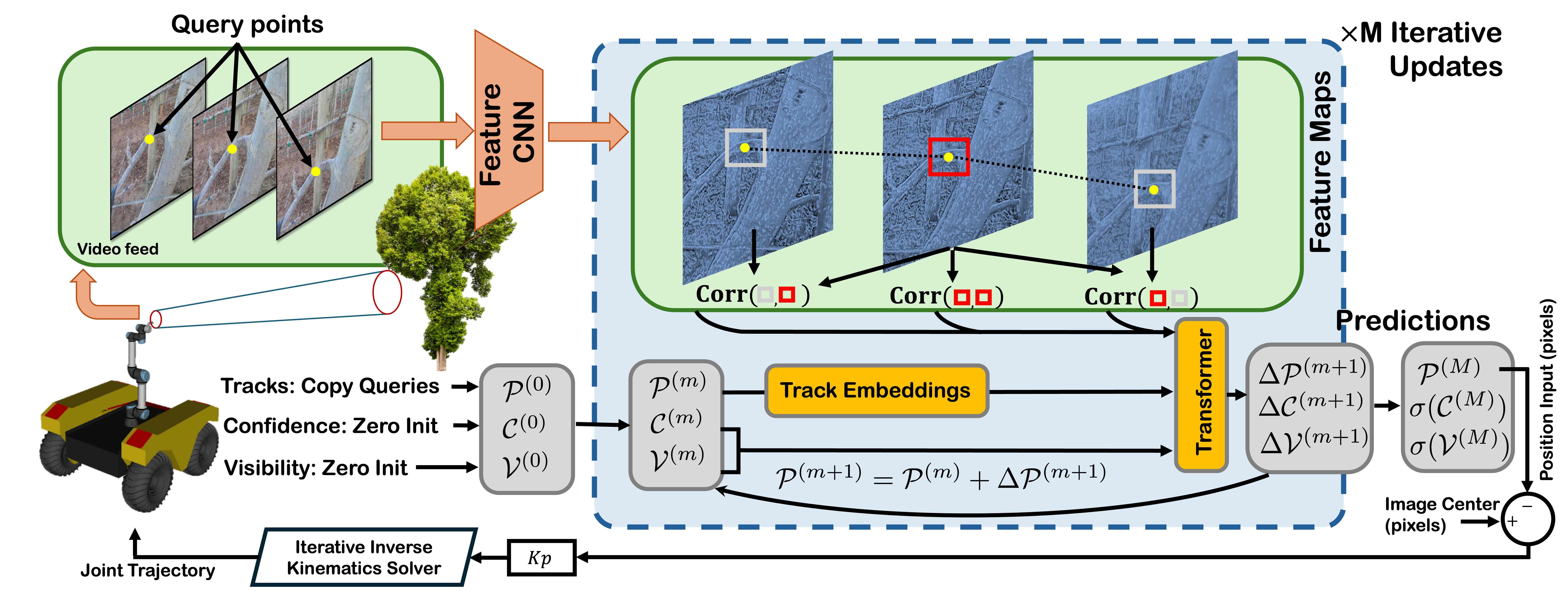}
    \caption{System Architecture: The tracking system uses a camera mounted on the robotic arm to capture visual data and feed to CoTracker3's perception pipeline \citep{karaev2024cotracker3} to extract convolutional features from each frame. The system analyzes feature correlations between frames to track the pruning point. A transformer iteratively refines the pruning point’s track, confidence, and visibility using previous estimates for accurate tracking.}
    \label{fig:2}
\end{figure*}
This section outlines this study's approach to precision pruning of orchard trees. The Gazebo simulation platform, integrated with the Robot Operating System (ROS) is used to create a controlled testing environment, as shown in Figure \ref{fig:1}. The robotic system consists of a Universal Robots UR5e manipulator mounted on a Clearpath Warthog mobile platform, with an Intel RealSense D435 camera attached to the manipulator's flange. The system's primary goal is to visually servo an end-effector toward a target branch, ensuring precise end-effector alignment with the selected pruning point.

Gazebo serves as the core simulation environment, providing a physics-based framework to model orchard trees, the robotic manipulator, mobile platform, and the surrounding workspace. The simulated orchard features dormant trees arranged in a tall spindle architecture, representing high-density planting systems commonly used in modern commercial orchards. Gazebo enables rendering of object interactions and collision dynamics, though the simulated sensor feedback may differ from real-world conditions. The simulation does not fully account for sensor noise, lighting variations, tree architecture differences and environmental complexities present in field conditions. Despite these limitations, Gazebo's high-fidelity physics engine and seamless compatibility with ROS provide a valuable platform for initial validation before real-world deployment \citep{koenig2004design}.

The vision controller utilizes CoTracker3 to track pruning points across video frames. Currently, pruning points are selected manually, but ongoing research explores AI-driven methods for automatic pruning point detection based on tree geometry. These advancements will be integrated into the system in future iterations. The system operates primarily in image space for visual servoing control, with depth information used solely as a stopping mechanism when the robot approaches within 20cm of the tracked pruning point. The following sections detail the key components of this study's methodology, including control architecture, feature tracking, controller implementation, and performance evaluation.

\subsection{CoTracker3}

CoTracker3 is a transformer-based point tracking model (see Figure \ref{fig:2}) that leverages joint attention mechanisms to track multiple points simultaneously, enabling robust performance under challenging conditions such as occlusions \citep{karaev2024cotracker}. Unlike traditional methods that track points independently, CoTracker3 can infer occluded point positions using information from visible points and prior frames, significantly improving tracking robustness in cluttered environments.

The model begins with user selection of an initial point $(x_0, y_0)$ which serves as a query embedding for tracking initialization. The system monitors depth and automatically stops visual servoing when the tracked point reaches within 20cm of the camera as a proximity-based termination condition.

For visual servoing control, the system computes an error vector between the tracked point $(x_t, y_t)$ and the image center $(x_c, y_c)$:
\begin{equation}
\mathbf{e}_t = \begin{bmatrix} x_c - x_t \\ y_c - y_t \end{bmatrix}
\end{equation}
This error vector drives the visual servoing control law, guiding the robotic manipulator to center the pruning point in the camera view for precise end-effector alignment.

\subsection{Image-Based Visual Servoing Controller}

We adopt a proportional (P) control strategy that operates directly on the 2D image-space error provided by CoTracker3. The proportional controller generates incremental motion commands in the image plane according to
\begin{equation}
\Delta x \;=\; K_p^x \cdot e_x,
\quad
\Delta z \;=\; K_p^z \cdot e_y.
\end{equation}
where $K_p^x,$ and $K_p^z \in \mathbb{R_{>\textrm{0}}}$ are empirically tuned positive proportional gains that map pixel errors to metric displacements in the robot frame. In our notation, $\mathbb{R_{>\textrm{0}}}$ denotes a positive real number. Horizontal image errors ($e_x$) map to lateral end-effector motions ($\Delta x$), while vertical image errors ($e_y$) map to vertical end-effector motions ($\Delta z$) as established by the camera mounting configuration. These conservative gain values ensure stable convergence without overshoot while maintaining responsive tracking performance. The error terms $(x_c - x_t)$ and $(y_c - y_t)$ naturally encode the required motion direction: when the pruning point appears to the right of the image center ($x_t > x_c$), the negative error drives the end-effector leftwards to recenter the target. Additionally, the system applies a constant forward motion ($\Delta y$) increment per control cycle to steadily advance the end-effector toward the pruning target.

These Cartesian commands define how the end-effector should move in the camera frame to reduce the 2D tracking error. However, the manipulator must execute these displacements through joint-space motion. The next subsection details how these Cartesian commands are mapped to corresponding joint angle updates via a iterative inverse kinematics approach.

\subsection{Iterative Inverse Kinematics}

To translate the image-based control signals into robot joint motions, an iterative inverse kinematics (IK) approach with singularity handling and joint limit enforcement is employed. The approach iteratively solves for pose errors and updates joint angles until convergence.

The manipulator's kinematics are defined using the Denavit-Hartenberg (DH) convention with parameters specific to the UR5e robot. Forward kinematics map joint angles $\mathbf{q} \in \mathbb{R}^6$ to end-effector poses in 3D space via a transformation matrix $\mathbf{T}_\mathrm{end}(\mathbf{q}) \in \mathbb{R}^{4\times 4}$, which encodes the end-effector pose in 3D space.

At each control iteration, the desired end-effector pose is computed by applying the visual servoing corrections to the current pose. The desired position incorporates the lateral correction $\Delta x$, forward motion $\Delta y$ and vertical correction $\Delta z$ while maintaining the current orientation to preserve end-effector alignment.

The IK solver iteratively minimizes the pose error between current and desired poses:
\begin{equation}
    \mathbf{e}_{\text{pose}} = \begin{bmatrix} \mathbf{e}_{\text{pos}} \\ \mathbf{e}_{\text{orient}} \end{bmatrix} = \begin{bmatrix} \mathbf{p}_{\text{target}} - \mathbf{p}_{\text{current}} \\ \text{Log}_{SO(3)}\left( \mathbf{R}_{\text{target}} \mathbf{R}_{\text{current}}^T \right) \end{bmatrix}
\end{equation}
where $\mathbf{e}_{\text{pos}}, \mathbf{e}_{\text{orient}} \in \mathbb{R}^3$ are the position and orientation errors respectively, and $\mathbf{R}_{\text{target}}, \mathbf{R}_{\text{current}} \in \mathbb{R}^{3 \times 3}$ are the target and current rotation matrices.

To address singularities and ensure numerical stability, we employ a damped pseudo-inverse formulation:
\begin{equation}
    \Delta\mathbf{q} = \mathbf{J}^T(\mathbf{J}\mathbf{J}^T + \lambda\mathbf{I})^{-1} \mathbf{e}_{\text{pose}}
\end{equation}
where $\mathbf{J} \in \mathbb{R}^{6 \times 6}$ is the numerical Jacobian computed using finite differences, $\lambda = 10^{-4}$ is the damping factor, and $\mathbf{I} \in \mathbb{R}^{6 \times 6}$ is the identity matrix.

The algorithm incorporates error-dependent step sizing to regulate convergence:
\begin{equation}
    \alpha = \text{clip}\left(0.01 \cdot (1.0 + \|\mathbf{e}_{\text{pose}}\|), 0.001, 0.05\right)
\end{equation}

with step size reduction near singularities when $\det(\mathbf{J}\mathbf{J}^T) < 10^{-6}$. The joint angles are updated as $\mathbf{q} \leftarrow \mathbf{q} + \alpha \cdot \Delta\mathbf{q}$, with step magnitude limited to 0.1 radians per iteration to prevent discontinuous motion.

The iteration terminates when both position and orientation errors satisfy convergence criteria ($\|\mathbf{e}_{\text{pos}}\| < 10^{-4}$ m and $\|\mathbf{e}_{\text{orient}}\| < 10^{-3}$ rad) or after a maximum of 1000 iterations. This iterative formulation ensures stable convergence while enforcing joint limits and preventing excessive rotations, particularly for wrist joints where angular displacements are constrained to minimize unnecessary motion. The visual servoing process continues until depth monitoring detects that the tracked point is within 20cm of the camera, at which point, motion commands cease to prevent contact with the pruning target. Currently, the system relies solely on this depth-based termination criterion. Future field work will incorporate force sensing control for more sophisticated contact detection and control.

\subsection{Evaluation Metrics}

The performance of the vision-based controller was evaluated using several quantitative metrics across 40 experimental trials. The 3D positioning error, $E_{pos}$, was calculated as the euclidean distance between the current end-effector position and the target pruning point:

\begin{equation}
    E_{pos} = \sqrt{(x_\textrm{ee} - x_\textrm{target})^2 + (y_\textrm{ee} - y_\textrm{target})^2 + (z_\textrm{ee} - z_\textrm{target})^2}
\end{equation}

where $(x_\textrm{ee}, y_\textrm{ee}, z_\textrm{ee})$ denotes the end-effector position in Cartesian space, and $(x_\textrm{target}, y_\textrm{target}, z_\textrm{target})$ represents the 3D coordinates of the target pruning point. The mean and standard deviation of end-effector positioning errors are computed to assess overall system accuracy and consistency.

The pixel-space tracking error, $E_\textrm{pixel}$, quantifies the magnitude of the error vector in image coordinates:
\begin{equation}
    E_\textrm{pixel} = \sqrt{(x_c - x_t)^2 + (y_c - y_t)^2}
\end{equation}

where $(x_c, y_c)$ is the image center and $(x_t, y_t)$ is the tracked point location. The mean and standard deviation of pixel errors provide insight into tracking performance throughout the approach sequence.

Success rates are evaluated at two precision levels: the percentage of trials achieving end-effector positioning within 5 mm of the target pruning point, and the percentage of trials achieving positioning within 10 mm of the target. These metrics provide comprehensive assessment of the system's precision capabilities for different tolerance requirements in pruning applications.

\section{Results and Discussion}\label{section3}
\begin{figure} \centering \includegraphics[width=0.9\linewidth]{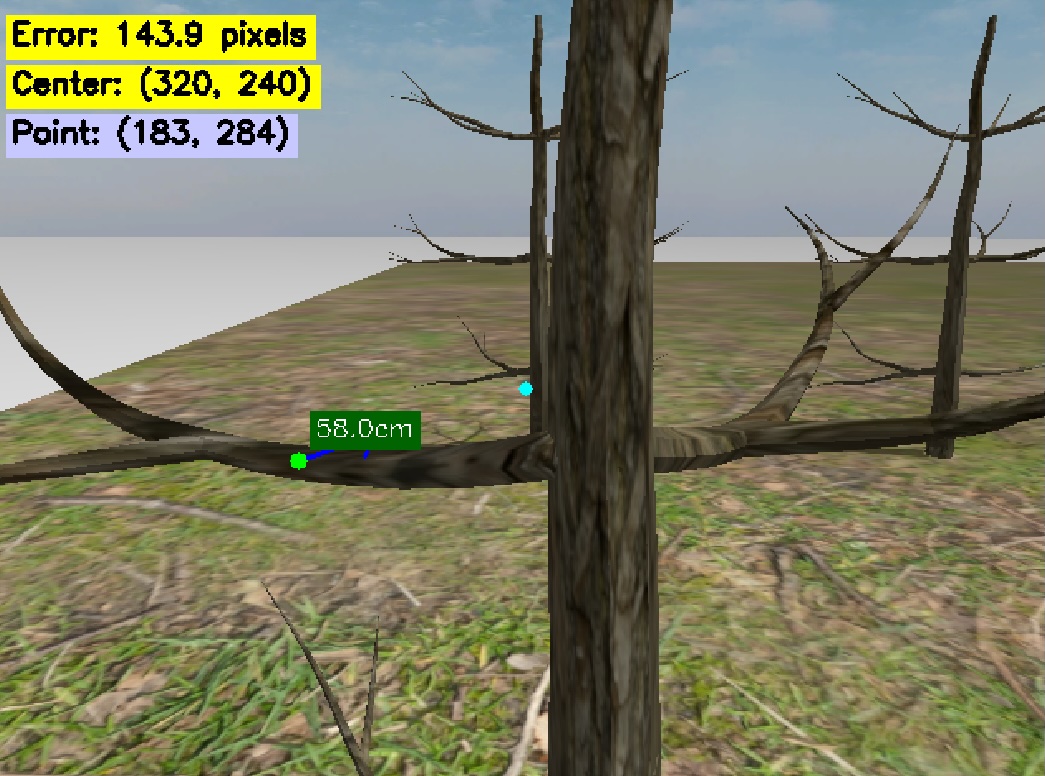} \caption{Simulated camera view of tracked point with statistics displayed. Green colored point indicates tracked pruning point which is to be centered  in image frame using UR5e joint-space motion.}\label{fig:camera_frame} \end{figure}

This section presents the experimental results and analysis of the vision-based controller's performance in orchard tree pruning tasks. The system was evaluated through 40 simulation trials conducted in Gazebo within a simulated orchard environment, with pruning points manually selected across the entire 640×480 pixel image plane at varying positions to assess the controller's robustness under diverse initial conditions. Figure \ref{fig:camera_frame} illustrates the simulation environment showing the camera view with tracked pruning points and the corresponding robot positioning in the Gazebo simulation. 
\begin{table}[h]
\centering
\caption{Summary of Quantitative Results from 40 Visual Servoing Trials}
\label{tab:results_summary}
\begin{tabular}{|c|c|c|c|}
\hline
Metric&Value\\
\hline
Total Trials & 40\\
Mean Pixel Error&9.79 px\\
Std Pixel Error&3.15 px\\
Mean EE Error&4.28 mm\\
Std EE error& 1.36 mm\\
Success Rate ($<5$ mm)&77.77\%\\
Success Rate ($<10$ mm)&100\%\\
\hline
\end{tabular}
\end{table}

Table~\ref{tab:results_summary} summarizes the quantitative performance metrics from the complete experimental dataset. The system achieved a mean end-effector positioning error of 4.28 ± 1.36 mm across all trials, demonstrating consistent sub-centimeter accuracy. Visual tracking performance showed a mean pixel error of 9.79 ± 3.15 pixels, indicating effective convergence of the proportional control scheme. Precision analysis reveals that 77.77\% of trials achieved positioning accuracy within 5 mm of the target, while 100\% of trials achieved accuracy within 10 mm, demonstrating the system's reliability for precision pruning applications.

\begin{figure} \centering \includegraphics[width=1.0\linewidth]{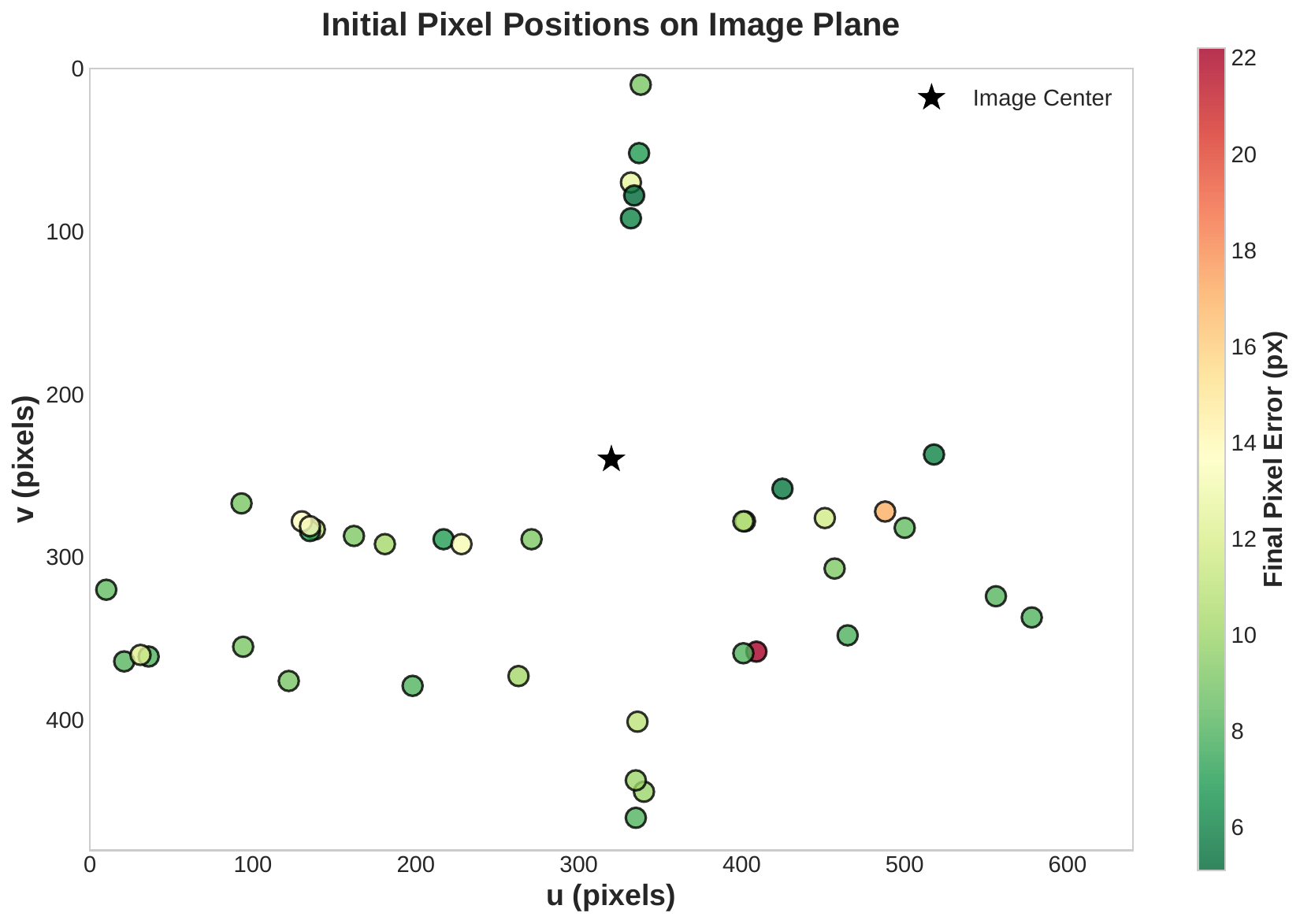} \caption{Spatial distribution of initial pixel positions colored by final tracking error}\label{fig:pixel_positions} \end{figure}

CoTracker3 demonstrated robust tracking performance across diverse target locations within the image plane. The algorithm successfully maintained target tracking from initial pixel offsets ranging from 50 to 350 pixels from the image center, with consistent convergence behavior regardless of starting position. Figure \ref{fig:pixel_positions} illustrates this spatial coverage, showing initial pixel positions color-coded by their final tracking error. The uniform error distribution across the workspace validated CoTracker3's reliability and the controller's ability to handle targets positioned anywhere within the camera field of view.

\begin{figure} \centering \includegraphics[width=1.0\linewidth]{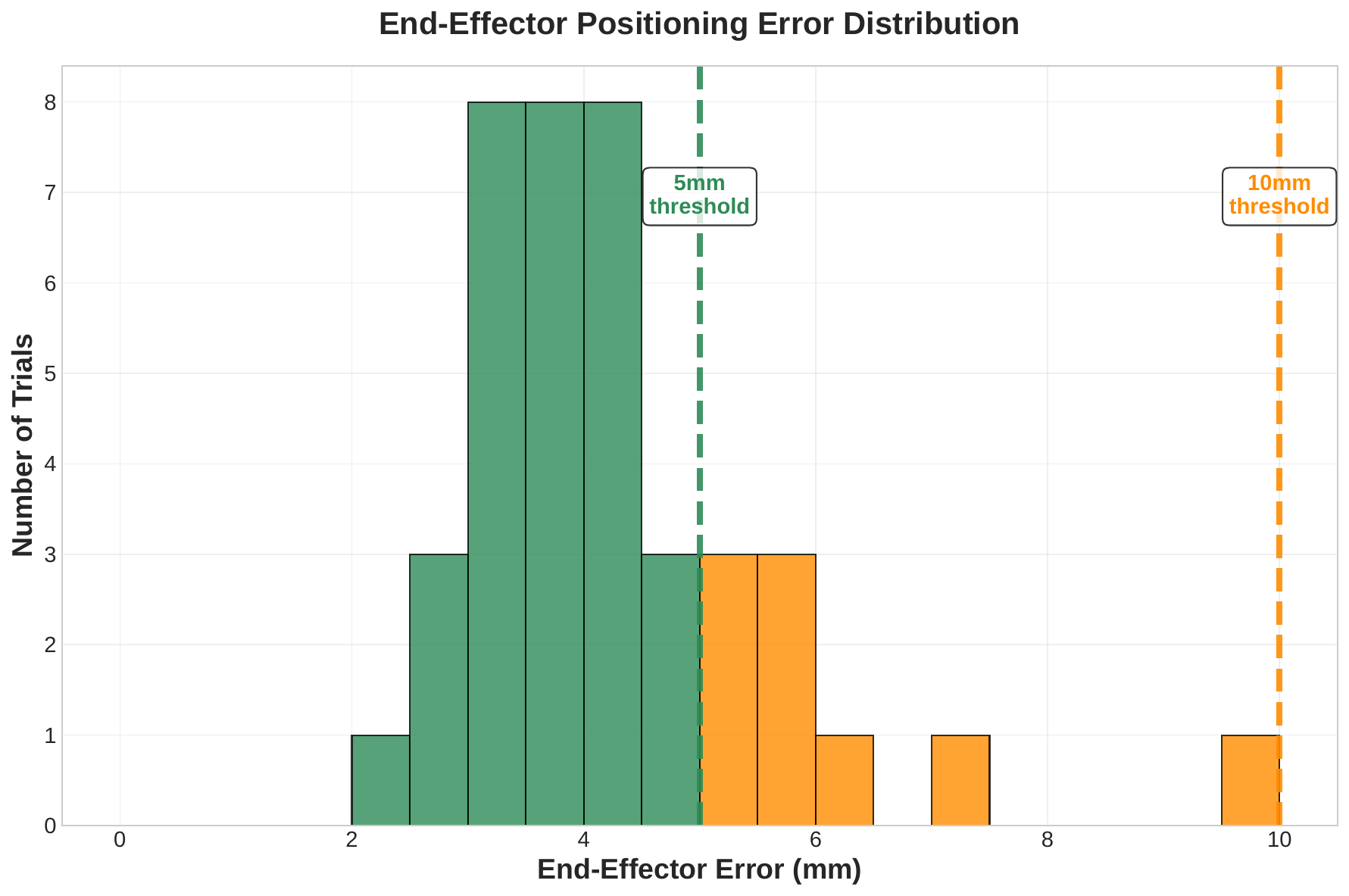} \caption{End-effector positioning error distribution with precision thresholds}\label{fig:error_dis} \end{figure}

The iterative inverse kinematics solver with damped pseudo-inverse formulation successfully handled all target configurations, achieving 100\% success rate within 10 mm tolerance. The adaptive step sizing and joint limit enforcement prevented singularities while maintaining smooth motion profiles throughout the approach sequences. Figure \ref{fig:error_dis} reveals the system's precision characteristics through the end-effector error distribution. Most trials achieved high precision with errors clustered below 5 mm, while the maximum error reached approximately 10 mm, well within acceptable tolerances for positioning applications. This distribution pattern confirmed the effectiveness of the integrated control approach and demonstrated consistent performance across all tested configurations.

The proportional gains ($Kp^x$, $Kp^z$) provided stable convergence without oscillatory behavior across all trials. The depth monitoring automatically halted tracking when the tracked point reached within 20 cm of the camera. The current implementation relies on 2D visual feedback with depth used solely to halt tracking, limiting precision to the achieved 9.79 pixel mean error. The results demonstrate consistent sub-centimeter accuracy across diverse target positions, validating the integrated approach for precision agricultural applications.

\section{Conclusion and Future Work}\label{section4}
\begin{figure}[ht]
    \centering
    \begin{subfigure}[b]{0.48\linewidth}
        \centering
        \includegraphics[width=\textwidth]{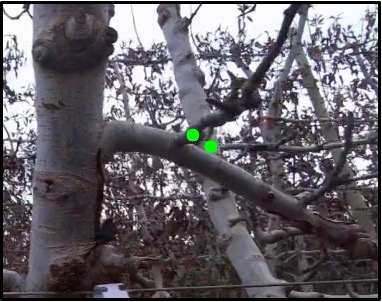}
        \caption{Frame at \( t_1 \)}
        \label{fig:occlusion_t1}
    \end{subfigure}
    \hfill
    \begin{subfigure}[b]{0.48\linewidth}
        \centering
        \includegraphics[width=\textwidth]{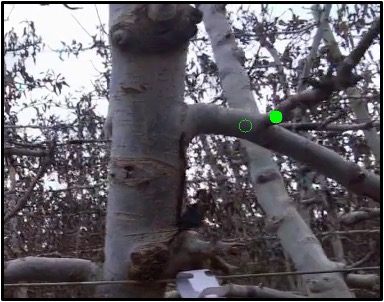}
        \caption{Frame at \( t_2 \)}
        \label{fig:occlusion_t2}
    \end{subfigure}

    \caption{Occlusion handling in CoTracker3. a) frame \( t_1 \) , the target is fully visible. b) In frame \( t_2 \), the target becomes occluded by another branch, yet tracking remains consistent, demonstrating CoTracker3's robustness in challenging orchard environments.}
    \label{fig:occlusion_handling}
\end{figure}

This study presented a vision-based robotic control system for precision tree pruning using CoTracker3 point tracking and iterative inverse kinematics. The system achieved 77.77\% accuracy within 5mm tolerance and 100\% within 10mm tolerance across 40 simulation trials, with a mean end-effector error of 4.28 ± 1.36 mm, demonstrating the effectiveness of integrating modern point tracking with classical control techniques for agricultural robotics. The instances where the controller failed to achieve sub-5mm precision were primarily attributed to point tracking error when the camera was driven too close to a branch.
To evaluate CoTracker3's robustness beyond simulation, the algorithm was tested on real-world imagery. Figure \ref{fig:occlusion_handling} exemplifies CoTracker3's occlusion handling capabilities in challenging orchard environments. In frame $t_1$, the target pruning point is clearly visible, while in frame $t_2$, the same point becomes occluded by overlapping branches. Despite this obstruction, CoTracker3 maintains consistent tracking through its joint attention mechanism, successfully predicting the target's location using spatiotemporal features from surrounding visible points.

While the current simulation trials operated in relatively unoccluded environments, this occlusion-handling capability will become critical in future work involving dense canopies, multiple simultaneous pruning points, and wind-induced branch movement. The demonstrated robustness positions CoTracker3 as an effective foundation for more complex agricultural scenarios where visual obstruction is unavoidable.
Future work will address several key limitations to enhance system capabilities. Future plans include integrating end-effector orientation control for angled cuts based on branch geometry, migrating from Gazebo to NVIDIA Isaac Sim for improved physics simulation and realistic force modeling, and incorporating force sensing for reliable contact detection and cut verification. Additionally, implementing collision avoidance algorithms and extending the framework to handle multiple simultaneous pruning points will significantly improve the system's practical applicability in commercial orchard operations.

\begin{ack}
This research is funded by the National Science Foundation and United States Department of Agriculture, National Institute of Food and Agriculture, United States through the following grants: AWD003473 as part of the “AI Institute for Agriculture (AgAID)” Program and 2023-67021-38908 under the International Collaboration Grant.
\end{ack}

\bibliography{ifacconf}             
                                                   







\end{document}